\documentclass[11pt,a4paper]{article} 
\usepackage[hyperref]{style/acl2019}
\usepackage{times}
\usepackage{latexsym}
\usepackage{float}
\usepackage{algorithm}
\usepackage{algpseudocode}
\usepackage{enumitem}
\usepackage{amsmath}
\usepackage{amssymb}
\usepackage{textcomp}
\usepackage{booktabs}
\usepackage{array}
\usepackage{bbm}
\usepackage{url}
\usepackage{svg}
\usepackage{xcolor}
\usepackage{soul}
\usepackage{seqsplit}
\usepackage{multirow}
\usepackage{adjustbox}
\usepackage{makecell}
\definecolor{mycolor}{HTML}{FD08FF}
\definecolor{mycolor1}{HTML}{0000AA}
\definecolor{mycolor2}{HTML}{BB0000}

\title{Syntactically Supervised Transformers\\ for Faster Neural Machine Translation}

\author{Nader Akoury, 
Kalpesh Krishna,
Mohit Iyyer \\
College of Information and Computer Sciences \\
University of Massachusetts Amherst \\
\texttt{\{nsa,kalpesh,miyyer\}@cs.umass.edu} \\
}

\aclfinalcopy
\newif\ifcomment 

\usepackage{mdwlist}
\usepackage{latexsym}
\usepackage{nicefrac}
\usepackage{booktabs}
\usepackage{amsfonts}
\usepackage{bold-extra}
\usepackage{amsmath}
\usepackage{dsfont}
\usepackage{amssymb}
\usepackage{bm}
\usepackage{graphicx}
\usepackage{mathtools}
\usepackage{microtype}
\usepackage{multirow}
\usepackage{multicol}
\usepackage{latexsym,comment}

\ifcomment
\newcommand{\micomment}[1]{\textcolor{red}{\bf \small [ #1 --Mohit]}}
\newcommand{\kkcomment}[1]{\textcolor{blue}{\bf \small [ #1 --Kalpesh]}}
\newcommand{\tvcomment}[1]{\textcolor{green}{\bf \small [ #1 --Tu]}}
\newcommand{\nacomment}[1]{\textcolor{orange}{\bf \small [ #1 --Nader]}}
\newcommand{\swcomment}[1]{\textcolor{yellow}{\bf \small [ #1 --Shufan]}}
\else
\newcommand{\micomment}[1]{ }
\newcommand{\tvcomment}[1]{ }
\newcommand{\kkcomment}[1]{ }
\newcommand{\nacomment}[1]{ }
\newcommand{\swcomment}[1]{ }
\fi

\newcommand{\gem}[1]{\mbox{\textsc{gem}}}

\newcommand{\hidetext}[1]{}
\newcommand{\ignore}[1]{}

\newcommand{\smallurl}[1]{ \begin{tiny}\url{#1}\end{tiny}}

\definecolor{lightblue}{HTML}{3cc7ea}
\definecolor{grey}{rgb}{0.95,0.95,0.95}
\definecolor{ceil}{rgb}{0.57, 0.63, 0.81}

\ifcomment
\newcommand{\mnicomment}[1]{  \colorbox{green}{   \parbox{.8\linewidth}{ MI: #1}  }}
\newcommand{\nsacomment}[1]{  \colorbox{yellow}{  \parbox{.8\linewidth}{ NA: #1}  }}
\else
\newcommand{\mnicomment}[1]{ }
\newcommand{\nsacomment}[1]{ }
\fi

\newcommand{\bvec}[1]{\boldsymbol{#1}}
\newcommand{\name}[0]{SynST}
\newcommand{\lt}[0]{LT}
\newcommand{\namedref}[2]{\hyperref[#2]{#1~\ref*{#2}}}

\newcommand{\tableref}[1]{\namedref{Table}{tab:#1}}

\newcommand{\boldblue}[1]{\textbf{\textcolor{mycolor1}{#1}}}
\newcommand{\itred}[1]{\textbf{\textcolor{mycolor2}{#1}}}
\newcommand{\invisiblebox}{\colorbox{white}{\phantom{Gy}}}

\setitemize{itemsep=2pt,topsep=2pt,parsep=0pt,partopsep=5pt}
\setenumerate{itemsep=2pt,topsep=2pt,parsep=0pt,partopsep=5pt,itemindent=13pt}

\begin{document}
\maketitle

\begin{abstract}
Standard decoders for neural machine translation \emph{autoregressively} generate a single target token per time step, which slows inference especially for long outputs. While architectural advances such as the Transformer fully parallelize the decoder computations at training time, inference still proceeds sequentially. Recent developments in \emph{non-} and \emph{semi-autoregressive} decoding produce multiple tokens per time step independently of the others, which improves inference speed but deteriorates translation quality. In this work, we propose the syntactically supervised Transformer (\name), which first autoregressively predicts a chunked parse tree before generating all of the target tokens in one shot conditioned on the predicted parse. A series of controlled experiments demonstrates that \name\ decodes sentences $\sim5\times$ faster than the baseline autoregressive Transformer while achieving higher BLEU scores than most competing methods on En-De and En-Fr datasets.
\end{abstract}

\section{Introduction}

Most models for neural machine translation (NMT) rely on \emph{autoregressive} decoders, which predict each token $t_i$ in the target language one by one conditioned on all previously-generated target tokens $t_{1\cdots i-1}$ and the source sentence $s$. For downstream applications of NMT that prioritize low latency (e.g., real-time translation), autoregressive decoding proves expensive, as decoding time in state-of-the-art attentional models such as the Transformer~\citep{transformer} scales quadratically with the number of target tokens.

\begin{figure}[t!]
\centering
\includegraphics[scale=0.57]{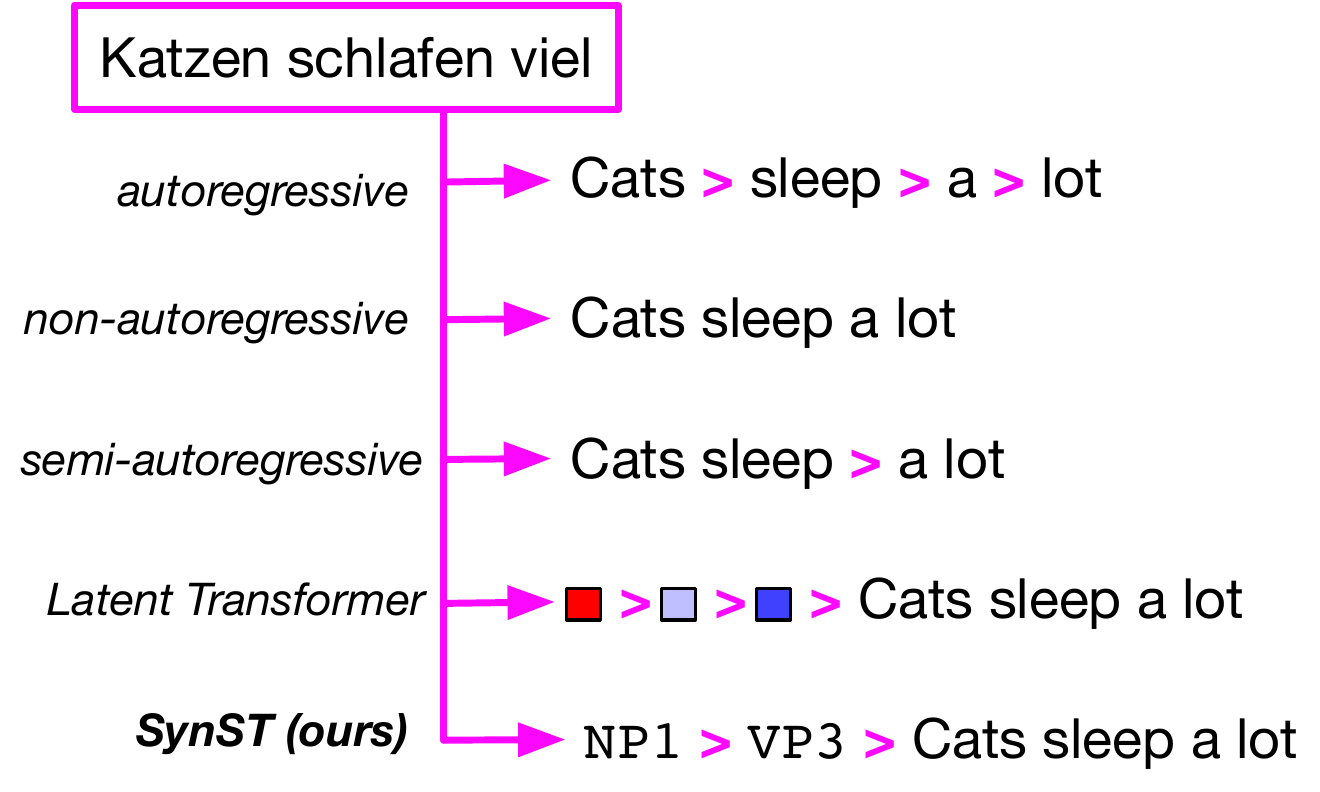}
\caption{Comparison of different methods designed to increase decoding speed. The arrow $\textcolor{mycolor}{\boldsymbol{>}}$ indicates the beginning of a new decode step conditioned on everything that came previously. The latent Transformer produces a sequence of discrete latent variables, whereas \name\ produces a sequence of syntactic constituent identifiers.}
\label{fig:overview}
\end{figure}

In order to speed up test-time translation, \emph{non-autoregressive} decoding methods produce all target tokens at once independently of each other ~\citep{nonAutoregressiveTransformer,lee2018deterministic}, while \emph{semi-autoregressive} decoding~\citep{semiAutoregressiveTransformer,parallelDecoding} trades off speed for quality by reducing (but not completely eliminating) the number of sequential computations in the decoder (Figure~\ref{fig:overview}). We choose the latent Transformer (\lt) of~\newcite{latentTransformer} as a starting point, which merges both of these approaches by autoregressively generating a short sequence of discrete latent variables before non-autoregressively producing all target tokens conditioned on the generated latent sequence.

\newcite{latentTransformer} experiment with increasingly complex ways of learning their discrete latent space, some of which obtain small BLEU improvements over a purely non-autoregressive baseline with similar decoding speedups. In this work, we propose to syntactically supervise the latent space, which results in a simpler model that produces better and faster translations.\footnote{Source code to reproduce our results is available at \url{https://github.com/dojoteef/synst}} Our model, the syntactically supervised Transformer (\name, Section~\ref{sec:model}), first autoregressively predicts a sequence of target syntactic chunks, and then non-autoregressively generates all of the target tokens conditioned on the predicted chunk sequence. During training, the chunks are derived from the output of an external constituency parser. We propose a simple algorithm on top of these parses that allows us to control the average chunk size, which in turn limits the number of autoregressive decoding steps we have to perform.

\name\ improves on the published \lt\ results for WMT 2014 En$\rightarrow$De in terms of both BLEU (20.7 vs. 19.8) and decoding speed ($4.9\times$ speedup vs. $3.9\times$). While we replicate the setup of~\newcite{latentTransformer} to the best of our ability, other work in this area does not adhere to the same set of datasets, base models, or ``training tricks'', so a legitimate comparison with published results is difficult. For a more rigorous comparison, we re-implement another related model within our framework, the semi-autoregressive transformer (SAT) of~\newcite{semiAutoregressiveTransformer}, and observe improvements in BLEU and decoding speed on both En$\leftrightarrow$ De and En$\rightarrow$ Fr language pairs (Section~\ref{sec:experiments}).

While we build on a rich line of work that integrates syntax into both NMT~\citep{parseGuidedNMT,eriguchi2017learning} and other language processing tasks~\citep{emnlp18-strubell,swayamdipta2018syntactic}, we aim to use syntax to speed up decoding, not improve downstream performance (i.e., translation quality). An in-depth analysis (Section~\ref{sec:analysis}) reveals that syntax is a powerful abstraction for non-autoregressive translation: for example, removing information about the constituent type of each chunk results in a drop of 15 BLEU on IWSLT En$\rightarrow$De. 

\begin{figure*}[t!]
\centering
\includegraphics[scale=0.58]{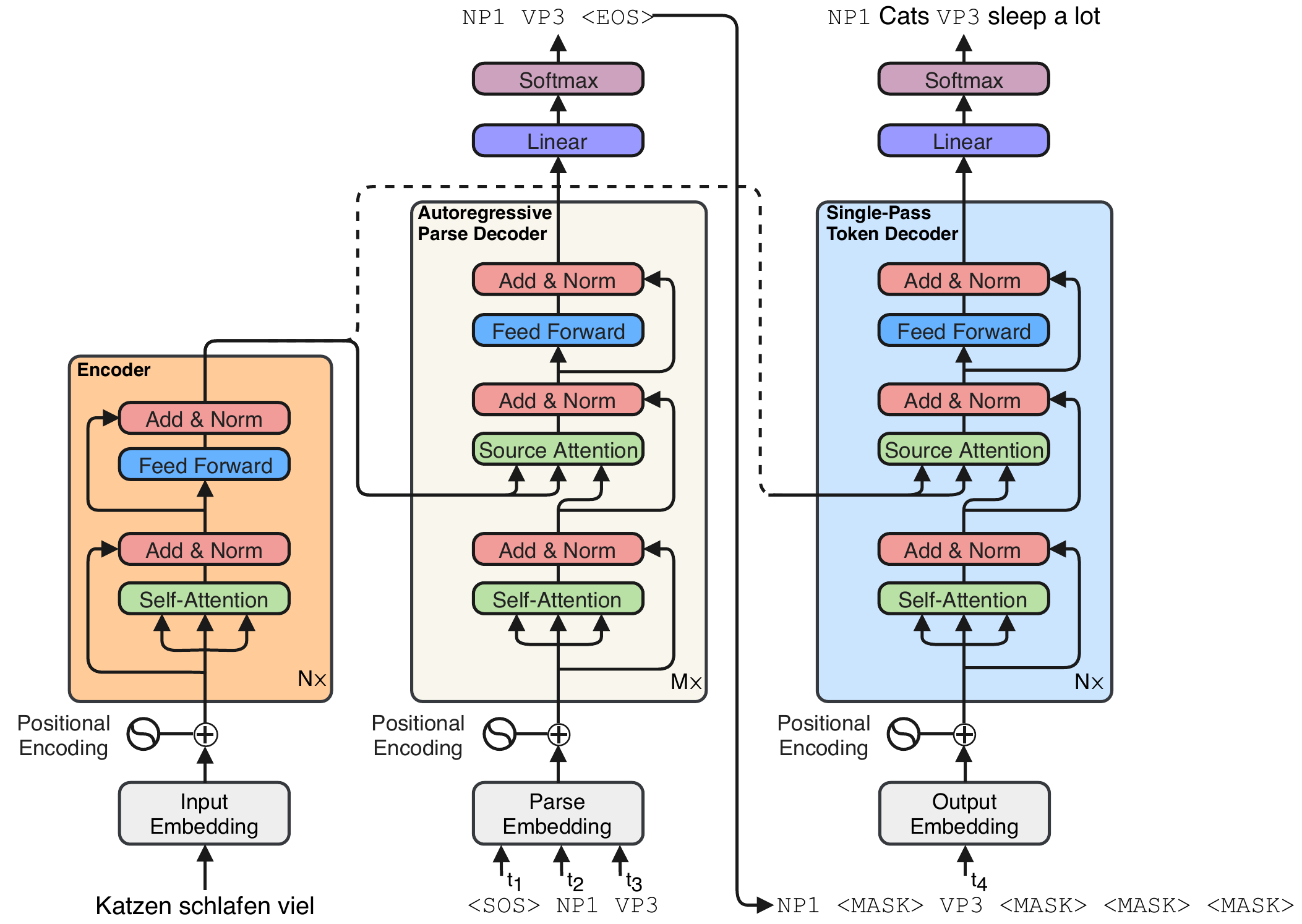}
\caption{A high-level overview of the \name\ architecture. During training, the parse decoder learns to autoregressively predict all chunk identifiers in parallel (time steps $t_{1,2,3}$), while the token decoder conditions on the ``ground truth'' chunk identifiers to predict the target tokens in one shot (time step $t_4$). During inference (shown here), the token decoder conditions on the autoregressively predicted chunk identifiers. The encoder and token decoder contain $N\geq 1$ layers, while the parse decoder only requires $M=1$ layers (see Table~\ref{tab:iwslt_bleu_speedup_tradeoff}).}
\label{fig:ourmodel}
\end{figure*}

\section{Decoding in Transformers}
\label{sec:background}

Our work extends the Transformer architecture ~\cite{transformer}, which is an instance of the encoder-decoder framework for language generation that uses stacked layers of self-attention to both encode a source sentence and decode the corresponding target sequence. In this section, we briefly review\footnote{We omit several architectural details in our overview, which can be found in full in~\newcite{transformer}.} the essential components of the Transformer architecture before stepping through the decoding process in both the vanilla autoregressive Transformer and non- and semi-autoregressive extensions of the model. 

\subsection{Transformers for NMT}
\label{subsec:transformer}
The Transformer encoder takes a sequence of source word embeddings $\bvec{s_{1,\cdots,}s_n}$ as input and passes it through multiple blocks of self-attention and feed-forward layers to finally produce contextualized token representations $\bvec{e_{1,\cdots,}e_n}$. Unlike recurrent architectures~\citep{hochreiter1997long,bahdanau2014neural}, the computation of $\bvec{e_n}$ does not depend on $\bvec{e_{n-1}}$, which enables full parallelization of the encoder's computations at both training and inference. To retain information about the order of the input words, the Transformer also includes positional encodings, which are added to the source word embeddings.

The decoder of the Transformer operates very similarly to the encoder during training: it takes a \emph{shifted} sequence of target word embeddings $\bvec{t_{1,\cdots,}t_m}$ as input and produces contextualized token representations $\bvec{d_{1,\cdots,}d_m}$, from which the target tokens are predicted by a softmax layer. Unlike the encoder, each block of the decoder also performs \emph{source attention} over the representations $\bvec{e_{1\dots n}}$ produced by the encoder. Another difference during training time is \emph{target-side masking}: at position $i$, the decoder's self attention should not be able to look at the representations of later positions $i+1,\ldots ,i+m$, as otherwise predicting the next token becomes trivial. To impose this constraint, the self-attention can be masked by using a lower triangular matrix with ones below and along the diagonal.

\subsection{Autoregressive decoding}
While at training time, the decoder's computations can be parallelized using masked self-attention on the ground-truth target word embeddings, inference still proceeds token-by-token. Formally, the vanilla Transformer factorizes the probability of target tokens $t_{1,\cdots,}t_m$ conditioned on the source sentence $s$ into a product of token-level conditional probabilities using the chain rule,
\begin{equation*}
  p(t_{1, \cdots,} t_m|s) = \prod_{i=1}^m p(t_i|t_{1,\cdots,} t_{i-1}, s).
\end{equation*}
During inference, computing $\arg\max_t p(t|s)$ is intractable, which necessitates the use of approximate algorithms such as beam search. Decoding requires a separate decode step to generate each target token $t_i$; as each decode step involves a full pass through every block of the decoder, autoregressive decoding becomes time-consuming especially for longer target sequences in which there are more tokens to attend to at every block. 

\subsection{Generating multiple tokens per time step}
As decoding time is a function of the number of decoding time steps (and consequently the number of passes through the decoder), faster inference can be obtained using methods that reduce the number of time steps. In autoregressive decoding, the number of time steps is equal to the target sentence length $m$; the most extreme alternative is (naturally) \emph{non-autoregressive} decoding, which requires just a single time step by factorizing the target sequence probability as
\begin{equation*}
  p(t_{1, \cdots,} t_m|s) = \prod_{i=1}^m p(t_i| s).
\end{equation*}
Here, all target tokens are produced \emph{independently} of each other. While this formulation does indeed provide significant decoding speedups, translation quality suffers after dropping the dependencies between target tokens without additional expensive  reranking steps~\citep[][NAT]{nonAutoregressiveTransformer} or iterative refinement with multiple decoders~\citep{lee2018deterministic}.

As fully non-autoregressive decoding results in poor translation quality, another class of methods produce $k$ tokens at a single time step where $1 < k < m$. The \emph{semi-autoregressive Transformer} (SAT) of~\newcite{semiAutoregressiveTransformer} produces a fixed $k$ tokens per time step, thus modifying the target sequence probability to:
\begin{equation*}
  p(t_{1, \cdots,} t_m|s) = \prod_{i=1}^{|G|} p(G_t|G_{1,\cdots,} G_{t-1}, x),
\end{equation*}
where each of $G_{1,\cdots,} G_{\left \lfloor{\frac{m-1}{k}}\right \rfloor + 1}$ is a group of contiguous non-overlapping target tokens of the form $t_{i,\cdots,}t_{i+k}$. In conjunction with training techniques like knowledge distillation~\citep{kim2016sequence} and initialization with an autoregressive model, SATs maintain better translation quality than non-autoregressive approaches with competitive speedups.~\newcite{parallelDecoding} follow a similar approach but dynamically select a different $k$ at each step, which results in further quality improvements with a corresponding decrease in speed.

\subsection{Latent Transformer}
While current semi-autoregressive methods achieve both better quality and faster speedups than their non-autoregressive counterparts, largely due to the number of tricks required to train the latter, the theoretical speedup for non-autoregressive models is of course larger. The latent Transformer~\citep[][\lt]{latentTransformer} is similar to both of these lines of work: its decoder first autoregressively generates a sequence of discrete latent variables $\bvec{l_{1,\cdots,}l_{j}}$ and then non-autoregressively produces the entire target sentence $t_{i,\cdots,}t_m$ conditioned on the latent sequence. Two parameters control the magnitude of the speedup in this framework: the length of the latent sequence ($j$), and the size of the discrete latent space ($K$). 

The \lt\ is significantly more difficult to train than any of the previously-discussed models, as it requires passing the target sequence through what~\newcite{latentTransformer} term a \emph{discretization bottleneck} that must also maintain differentiability through the decoder. While \lt\ outperforms the NAT variant of non-autoregressive decoding in terms of BLEU, it takes longer to decode. In the next section, we describe how we use syntax to address the following three weaknesses of \lt: 

\begin{enumerate}
  \item generating the same number of latent variables $j$ regardless of the length of the source sentence, which hampers output quality
  \item relying on a large value of $K$ (the authors report that in the base configuration as few as $\sim$3000 latents are used out of $2^{16}$ available), which hurts translation speed
  \item the complexity of implementation and optimization of the discretization bottleneck, which negatively impacts both quality and speed.
\end{enumerate}

\section{Syntactically Supervised Transformers}
\label{sec:model}

Our key insight is that we can use syntactic information as a proxy to the learned discrete latent space of the \lt. Specifically, instead of producing a sequence of latent discrete variables, our model produces a sequence of phrasal chunks derived from a constituency parser. During training, the chunk sequence prediction task is supervised, which removes the need for a complicated discretization bottleneck and a fixed sequence length $j$. Additionally, our chunk vocabulary is much smaller than that of the \lt, which improves decoding speed.

Our model, the syntactically supervised Transformer (\name), follows the two-stage decoding setup of the \lt. First, an autoregressive decoder generates the phrasal chunk sequence, and then all of the target tokens are generated at once, conditioned on the chunks (Figure~\ref{fig:ourmodel}). The rest of this section fully specifies each of these two stages.

\begin{figure}[t!]
\centering
\includegraphics[scale=0.34]{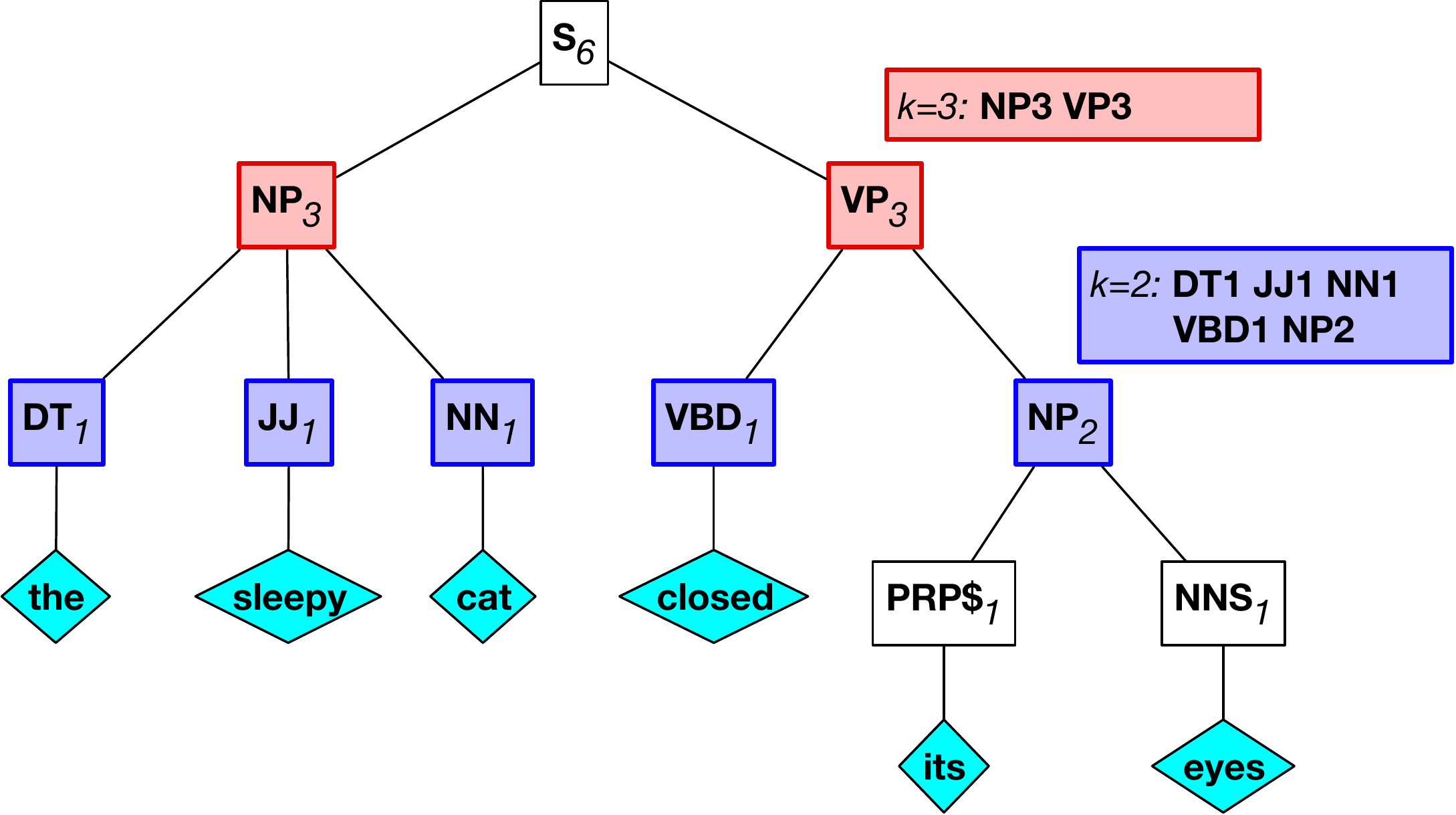}
\caption{Example of our parse chunk algorithm with max span sizes $k=2,3$. At each visited node during an in-order traversal of the parse, if the subtree size is less than or equal to $k$, we append a corresponding chunk identifier to our sequence.}
\label{fig:parse_alg}
\end{figure}

\subsection{Autoregressive chunk decoding}
Intuitively, our model uses syntax as a scaffold for the generated target sentence. During training, we acquire supervision for the syntactic prediction task through an external parser in the target language. While we could simply force the model to predict the entire linearized parse minus the terminals,\footnote{This approach is used for paraphrase generation by~\newcite{IyyerSCPN2018}, who were not focused on decoding speed.} this approach would dramatically increase the number of autoregressive steps, which we want to keep at a minimum to prioritize speed. To balance syntactic expressivity with the number of decoding time steps, we apply a simple chunking algorithm to the constituency parse.\bigbreak

\noindent\textbf{Extracting chunk sequences:} Similar to the SAT method, we first choose a maximum chunk size $k$. Then, for every target sentence in the training data, we  perform an in-order traversal of its constituency parse tree. At each visited node, if the number of leaves spanned by that node is less than or equal to $k$, we append a descriptive \emph{chunk identifier} to the parse sequence before moving onto its sibling; otherwise, we proceed to the left child and try again. This process is shown for two different values of $k$ on the same sentence in Figure~\ref{fig:parse_alg}. Each unique chunk identifier, which is formed by the concatenation of the constituent type and subtree size (e.g., \texttt{NP3}), is considered as an element of our first decoder's vocabulary; thus, the maximum size of this vocabulary is $|P|\times k$ where $P$ is the set of all unique constituent types.\footnote{In practice, this vocabulary is significantly smaller than the discrete latent space of the \lt\ for reasonable values of $k$.}
 Both parts of the chunk identifier (the constituent type and its size) are crucial to the performance of \name, as demonstrated by the ablations in Section~\ref{sec:analysis}.\bigbreak

\noindent\textbf{Predicting chunk sequences:}
Because we are fully supervising the chunk sequence prediction, both the encoder and parse decoder are architecturally identical to the encoder and decoder of the vanilla Transformer, respectively. The parse decoder differs in its target vocabulary, which is made up of chunk identifiers instead of word types, and in the number of layers (we use 1 layer instead of 6, as we observe diminishing returns from bigger parse decoders as shown in Section~\ref{sec:analysis}). Formally, the parse decoder autoregressively predicts a sequence of chunk identifiers $c_{1,\cdots,}c_p$ conditioned on the source sentence $s$\footnote{In preliminary experiments, we also tried conditioning this decoder on the source parse, but we did not notice significant differences in translation quality.} by modeling
\begin{equation*}
  p(c_{1, \cdots,} c_p|s) = \prod_{i=1}^p p(c_i|c_{1, \cdots,} c_{i-1}, s).
\end{equation*}
Unlike \lt, the length $p$ of the chunk sequence changes dynamically based on the length of the target sentence, which is reminiscent of the token decoding process in the SAT.

\subsection{Non-autoregressive token decoding}
In the second phase of decoding, we apply a single non-autoregressive step to produce the tokens of the target sentence by factorizing the target sequence probability as
\begin{equation*}
  p(t_{1, \cdots,} t_m|s) = \prod_{i=1}^m p(t_i|c_{1, \cdots,} c_p, s).
\end{equation*}
Here, all target tokens are produced independently of each other, but in contrast to the previously-described non-autoregressive models, we additionally condition each prediction on the entire chunk sequence. To implement this decoding step, we feed a chunk sequence as input to a second Transformer decoder, whose parameters are separate from those of the parse decoder. During training, we use the ground-truth chunk sequence as input, while at inference we use the predicted chunks.\bigbreak

\noindent\textbf{Implementation details:}
To ensure that the number of input and output tokens in the second decoder are equal, which is a requirement of the Transformer decoder, we add placeholder \texttt{<MASK>} tokens to the chunk sequence, using the size component of each chunk identifier to determine where to place these tokens. For example, if the first decoder produces the chunk sequence \texttt{NP2 PP3}, our second decoder's input becomes \texttt{NP2} \texttt{<MASK>} \texttt{<MASK>} \texttt{PP3} \texttt{<MASK>} \texttt{<MASK>} \texttt{<MASK>}; this formulation also allows us to better leverage the Transformer's positional encodings. Then, we apply unmasked self-attention over this input sequence and predict target language tokens at each position associated with a \texttt{<MASK>} token.

\section{Experiments}
\label{sec:experiments}
\newcounter{savedfootnote}
\newcounter{savedHfootnote}
\setcounter{savedfootnote}{\value{footnote}}
\setcounter{savedHfootnote}{\value{Hfootnote}}

\begin{table*}[t]
\begin{center}
\begin{tabular}{ lcccccccc }
 \toprule
 \textbf{Model} & \multicolumn{2}{c}{\underline{\textbf{WMT En-De}}} & \multicolumn{2}{c}{\underline{\textbf{WMT De-En}}}  & \multicolumn{2}{c}{\underline{\textbf{IWSLT En-De}}} & \multicolumn{2}{c}{\underline{\textbf{WMT En-Fr}}}\\
 & BLEU & Speedup & BLEU & Speedup & BLEU & Speedup & BLEU & Speedup \\
 \midrule
 Baseline ($b=1$) & 25.82 & $1.15\times$ & 29.83 & $1.14\times$  & 28.66 & $1.16\times$ & 39.41 & $1.18\times$ \\
 Baseline ($b=4$) & 26.87 & $1.00\times$ & 30.73 & $1.00\times$  & 30.00 & $1.00\times$ & 40.22 & $1.00\times$ \\
 \hline
 SAT ($k=2$) & 22.81 & $2.05\times$ & 26.78 & $2.04\times$ & 25.48 & $2.03\times$ & 36.62 & $2.14\times$ \\
 SAT ($k=4$) & 16.44 & $3.61\times$ & 21.27 & $3.58\times$ & 20.25 & $3.45\times$ & 28.07 & $3.34\times$ \\
 SAT ($k=6$) & 12.55 & $4.86\times$ & 15.23 & $4.27\times$ & 14.02 & $4.39\times$ & 24.63 & $4.77\times$ \\
 \hline
 \lt*  & 19.8 & $3.89\times$ & - & - & - & - & - & - \\
 \hline
 \name ($k=6$) & 20.74 & $4.86\times$ & 25.50 & $5.06\times$ & 23.82 & $3.78\times$ & 33.47 & $5.32\times$ \\
 \bottomrule
\end{tabular}
\end{center}
\caption{Controlled experiments comparing \name\ to a baseline Transformer, SAT, and \lt\ on four different datasets (two language pairs) demonstrate speed and BLEU improvements. Wall-clock speedup is measured on a single Nvidia TitanX Pascal by computing the average time taken to decode a single sentence in the dev/test set, averaged over five runs. When beam width $b$ is not specified, we perform greedy decoding (i.e., $b=1$). Note that the LT results are reported by~\newcite{latentTransformer} and not from our own implementation;\addtocounter{footnote}{3}\addtocounter{Hfootnote}{3}\footnotemark\setcounter{footnote}{\value{savedfootnote}}\setcounter{Hfootnote}{\value{savedHfootnote}} as such, they are not directly comparable to the other results.}
\label{tab:experiments}
\end{table*}

We evaluate the translation quality (in terms of BLEU) and the decoding speedup (average time to decode a sentence) of \name\ compared to competing approaches. In a controlled series of experiments on four different datasets (En$\leftrightarrow$ De and En$\rightarrow$ Fr language pairs),\footnote{We explored translating to other languages previously evaluated in the non- and semi-autoregressive decoding literature, but could not find publicly-available, reliable constituency parsers for them.} we find that \name\ achieves a strong balance between quality and speed, consistently outperforming the semi-autoregressive SAT on all datasets and the similar \lt\ on the only translation dataset for which~\newcite{latentTransformer} report results. In this section, we first describe our experimental setup and its differences to those of previous work before providing a summary of the key results.

\subsection{Controlled experiments}
Existing papers in non- and semi-autoregressive approaches do not adhere to a standard set of datasets, base model architectures, training tricks, or even evaluation scripts. This unfortunate disparity in evaluation setups means that numbers between different papers are uncomparable, making it difficult for practitioners to decide which method to choose. In an effort to offer a more meaningful comparison, we strive to keep our experimental conditions as close to those of~\newcite{latentTransformer} as possible, as the \lt\ is the most similar existing model to ours. In doing so, we made the following decisions:

\begin{itemize}
  \item Our base model is the base vanilla Transformer~\citep{transformer} without any architectural upgrades.\footnote{As the popular Tensor2Tensor implementation is constantly being tweaked, we instead re-implement the Transformer as originally published and verify that its results closely match the published ones. Our implementation achieves a BLEU of 27.69 on WMT'14 En-De, when using \texttt{multi-bleu.perl} from Moses SMT.}
  \item We use all of the hyperparameter values from the original Transformer paper and do not attempt to tune them further, except for: (1) the number of layers in the parse decoder, (2) the decoders do not use label smoothing.
	\item We do not use sequence-level knowledge distillation, which augments the training data with translations produced by an external autoregressive model. The choice of  model used for distillation plays a part in the final BLEU score, so we remove this variable.
  \item We report all our BLEU numbers using sacreBLEU \cite{sacreBLEU} to ensure comparability with future work.\footnote{SacreBLEU signature: \seqsplit{BLEU+case.mixed+lang.LANG+numrefs.1+smooth.exp+test.TEST+tok.intl+version.1.2.11}, with LANG $\in\{$en-de, de-en, en-fr$\}$ and TEST $\in\{$wmt14/full, iwslt2017/tst2013$\}$}
  \item We report wall-clock speedups by measuring the average time to decode one sentence (batch size of one) in the dev/test set.
\end{itemize}

\begin{table*}[t]
	\begin{center}
		\small
    \begin{adjustbox}{max width=\textwidth}
		\begin{tabular}{ p{2cm}p{1.9cm}p{11cm} }

			\toprule

			 & \textbf{Chunk types} & \textbf{SynST predictions with $\vert$ separating syntax chunks} \\
			\cmidrule{2-3}
			Words repeated in two separate syntax chunks \newline (\boldblue{blue}, \itred{red}) & \colorbox{white}{\boldblue{\texttt{NP1}}, \itred{\texttt{NP3}}} \newline \colorbox{white}{\boldblue{\texttt{NP3}}, \itred{\texttt{PP4}}} \newline \colorbox{white}{\boldblue{\texttt{NP2}}, \itred{\texttt{PP3}}} & But $\vert$ it $\vert$ is $\vert$ \colorbox{white}{\boldblue{enthusiasm}} $\vert$ in $\vert$ \colorbox{white}{\itred{a great enthusiasm}} \newline ... Enrique $\vert$ Pena $\vert$ Nieto $\vert$ is $\vert$ facing $\vert$ \colorbox{white}{\boldblue{a difficult start}} $\vert$ \colorbox{white}{\itred{on a difficult start}} \newline Do $\vert$ you $\vert$ not $\vert$ turn $\vert$ \colorbox{white}{\boldblue{your voters}} $\vert$ \colorbox{white}{\itred{on your voters}}  \\

			\toprule

			& \textbf{Output type} & \textbf{Output for a single example} \\
			\cmidrule{2-3}

			SynST reorders \boldblue{syntax chunks}, which is fixed with gold parses (GP) as input & \colorbox{white}{ground truth} \newline \colorbox{white}{SynST} \newline \colorbox{white}{predicted parse} \colorbox{white}{SynST + GP} & Canada $\vert$ was $\vert$ the first country $\vert$ to  $\vert$ make $\vert$ photograph warnings $\vert$\colorbox{white}{\boldblue{mandatory in 2001}} \newline Canada $\vert$ was $\vert$\colorbox{white}{the first country}$\vert$ \boldblue{in 2001} $\vert$ to $\vert$ propose $\vert$ photographic warnings \newline \texttt{NP1} \texttt{VBD1} \texttt{NP3} \colorbox{white}{\boldblue{\texttt{PP2}}} \texttt{TO1} \texttt{VB1} \texttt{NP4} \invisiblebox \newline Canada $\vert$ was $\vert$ the first country $\vert$ to $\vert$ make $\vert$ photographic warnings  $\vert$\colorbox{white}{\boldblue{available in 2001}} \\

			\toprule

			& \textbf{True chunk} & \textbf{SynST predictions with $@@$ as subword divisions}\\
			\cmidrule{2-3}

			Wrong subword completion within a \boldblue{syntax chunk} & \colorbox{white}{ignores them} \newline \colorbox{white}{beforehand} \newline \colorbox{white}{examines} & I $\vert$ simply $\vert$ \colorbox{white}{\boldblue{ign$@@$ it them}} \newline Most ST$@@$ I $\vert$ can $\vert$ be $\vert$ cur$@@$ ed $\vert$ \colorbox{white}{\boldblue{be$@@$ foreh$@@$ ly}} \newline  Beg$@@$ inning $\vert$ of $\vert$ the course $\vert$  which $\vert$ \colorbox{white}{\boldblue{exam$@@$ ates}} $\vert$ the ...\\

			\bottomrule
		\end{tabular}
    \end{adjustbox}
	\end{center}
	\caption{Common error made by SynST due to its syntactically informed semi-autoregressive decoding. Different syntax chunks have been separated by $\vert$ symbols in all the decoded outputs. }
	\label{tab:errors}
\end{table*}

As the code for \lt\ is not readily available\footnote{We attempted to use the publicly available code in Tensor2Tensor, but were unable to successfully train a model.}, we also reimplement the SAT model using our setup, as it is the most similar model outside of \lt\ to our own.\footnote{The published SAT results use knowledge distillation and different hyperparameters than the vanilla Transformer, most notably a tenfold decrease in training steps due to initializing from a pre-trained Transformer.} For \name, we set the maximum chunk size $k=6$ and compare this model to the SAT trained with $k=2,4,6$.

\subsection{Datasets}
We experiment with English-German and English-French datasets, relying on constituency parsers in all three languages. We use the Stanford CoreNLP \cite{corenlp} shift-reduce parsers for English, German, and French. For English-German, we evaluate on WMT 2014 En$\leftrightarrow$De as well as IWSLT 2016 En$\rightarrow$De, while for English-French we train on the Europarl / Common Crawl subset of the full WMT 2014 En$\rightarrow$Fr data and evaluate over the full dev/test sets. WMT 2014 En$\leftrightarrow$De  consists of around 4.5 million sentence pairs encoded using byte pair encoding~\citep{bpe} with a shared source-target vocabulary of roughly 37000 tokens. We use the same preprocessed dataset used in the original Transformer paper and also by many subsequent papers that have investigated improving decoding speed, evaluating on the \texttt{newstest2013} dataset for validation and the \texttt{newstest2014} dataset for testing. For the IWSLT dataset we use \texttt{tst2013} for validation and utilize the same hyperparameters as~\newcite{lee2018deterministic}.

\subsection{Results}
Table~\ref{tab:experiments} contains the results on all four datasets. \name\ achieves speedups of $\sim 4-5\times$ that of the vanilla Transformer, which is larger than nearly all of the SAT configurations. Quality-wise, \name\ again significantly outperforms the SAT configurations at comparable speedups on all datasets. On WMT En-De, \name\ improves by 1 BLEU over \lt\ (20.74 vs \lt's 19.8 without reranking).\bigbreak

\noindent\textbf{Comparisons to other published work: } As mentioned earlier, we adopt a very strict set of experimental conditions to evaluate our work against \lt\ and SAT. For completeness, we also offer an unscientific comparison to other numbers in \tableref{experiments-uncontrolled}.

\section{Analysis}
\label{sec:analysis}

In this section, we perform several analysis and ablation experiments on the IWSLT En-De dev set to shed more light on how \name\ works. Specifically, we explore common classes of translation errors, important factors behind \name's speedup, and the performance of \name's parse decoder.

\begin{figure}[h]
\centering
\includegraphics[scale=0.5]{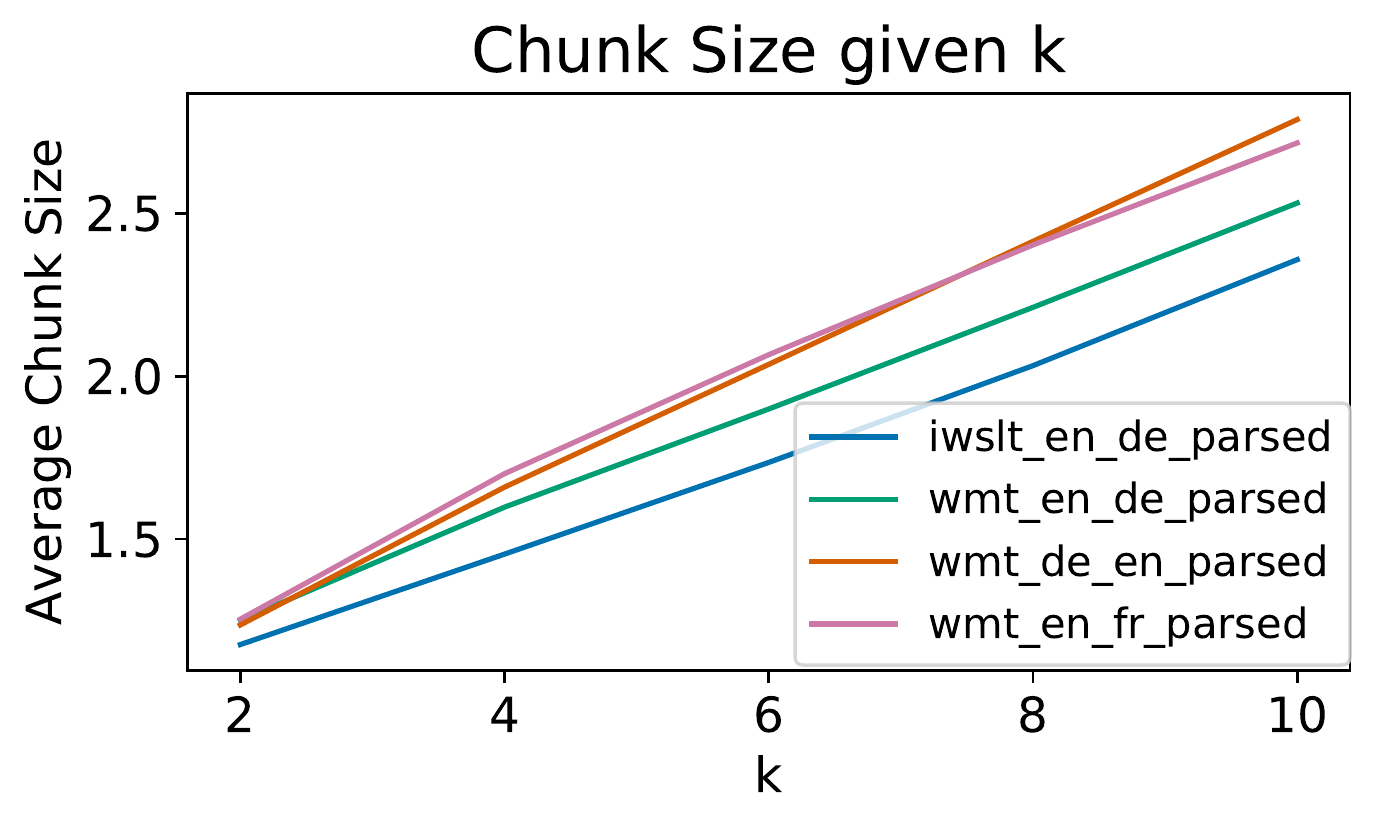}
\caption{The average size of a chunk given a particular value of the max chunk size $k$.}
\label{fig:constituent_span}
\end{figure}

\begin{table*}[t]
  \centering
  \begin{tabular}{lp{3cm}p{3cm}p{3cm}p{3cm}}
    \toprule &
    \footnotesize \bf \thead{Predicted parse\\vs.\\Gold parse (separate)} &
    \footnotesize \bf \thead{Predicted parse\\vs.\\Gold parse (joint)} &
    \footnotesize \bf \thead{Parsed prediction\\vs.\\Gold parse} &
    \footnotesize \bf \thead{Parsed prediction\\vs.\\Predicted parse} \\
    \midrule
    F1 & $65.48$ & $69.64$ & $79.16$ & $89.90$ \\
    Exact match & $4.23\%$ & $5.24\%$ & $5.94\%$ & $43.10\%$ \\
    \bottomrule
  \end{tabular}
  \caption{F1 and exact match comparisons of predicted chunk sequences (from the parse decoder), ground-truth chunk sequences (from an external parser in the target language), and chunk sequences obtained after parsing the translation produced by the token decoder. First two columns show the improvement obtained by jointly training the two decoders. The third column shows that when the token decoder deviates from the predicted chunk sequence, it usually results in a translation that is closer to the ground-truth target syntax, while the fourth column shows that the token decoder closely follows the predicted chunk sequence.}
  \label{tab:synst_parse_stats}
\end{table*}

\subsection{Analyzing \name's translation quality}

\noindent\emph{What types of translation errors does \name\ make? }
Through a qualitative inspection of \name's output translations, we identify three types of errors that \name\ makes more frequently than the vanilla Transformer: subword repetition, phrasal reordering, and inaccurate subword completions. Table~\ref{tab:errors} contains examples of each error type.\smallbreak

\noindent\emph{Do we need to include the constituent type in the chunk identifier? }\name's chunk identifiers contain both the constituent type as well as chunk size. Is the syntactic information actually useful during decoding, or is most of the benefit from the chunk size? To answer this question, we train a variant of \name\ without the constituent identifiers, so instead of predicting \texttt{NP3 VP2 PP4}, for example, the parse decoder would predict \texttt{3 2 4}. This model substantially underperforms, achieving a BLEU of 8.19 compared to 23.82 for \name, which indicates that the syntactic information is of considerable value.\smallbreak

\noindent\emph{How much does BLEU improve when we provide the ground-truth chunk sequence? } 
To get an upper bound on how much we can gain by improving \name's parse decoder, we replace the input to the second decoder with the ground-truth chunk sequence instead of the one generated by the parse decoder. The BLEU increases from 23.8 to 41.5 with this single change, indicating that future work on \name's parse decoder could prove very fruitful.\smallbreak

\subsection{Analyzing \name's speedup}
\noindent\emph{What is the impact of average chunk size on our measured speedup?}
Figure~\ref{fig:constituent_span} shows that the IWSLT dataset, for which we report the lowest \name\ speedup, has a significantly lower average chunk size than that of the other datasets at many different values of $k$.\footnote{IWSLT is composed of TED talk subtitles. A small average chunk size is likely due to including many short utterances.} We observe that our empirical speedup directly correlates with the average chunk size: ranking the datasets by empirical speedups in Table~\ref{tab:experiments} results in the same ordering as Figure~\ref{fig:constituent_span}'s ranking by average chunk size.\smallbreak

\noindent\emph{How does the number of layers in \name's parse decoder affect the BLEU/speedup tradeoff?} 
All \name\ experiments in Table~\ref{tab:experiments} use a single layer for the parse decoder. Table~\ref{tab:iwslt_bleu_speedup_tradeoff} shows that increasing the number of layers from 1 to 5 results in a BLEU increase of only 0.5, while the speedup drops from $3.8\times$ to $1.4\times$. Our experiments indicate that (1) a single layer parse decoder is reasonably sufficient to model the chunked sequence and (2) despite its small output vocabulary, the parse decoder is the bottleneck of SynST in terms of decoding speed.

\begin{table}[ht]
  \centering
  \begin{adjustbox}{max width=\columnwidth}
  \begin{tabular}{cccc}
    \toprule
    {\# Layers} & Max Chunk Size & Speedup & BLEU \\
    \midrule
    1 & $k=6$ & $3.8\times$ & 23.82 \\
    2 & $k=6$ & $2.8\times$ & 23.98 \\
    3 & $k=6$ & $2.2\times$ & 24.54 \\
    4 & $k=6$ & $1.8\times$ & 24.04 \\
    5 & $k=6$ & $1.4\times$ & 24.34 \\
    \midrule
    1 & $k\in\{1\ldots6\}$ & $3.1\times$ & 25.31 \\
    \bottomrule
  \end{tabular}
  \end{adjustbox}
  \caption{Increasing the number of layers in \name's parse decoder significantly lowers the speedup while marginally impacting BLEU. Randomly sampling $k$ from $\{1\ldots6\}$ during training boosts BLEU significantly with minimal impact on speedup.}
\label{tab:iwslt_bleu_speedup_tradeoff}
\end{table}

\subsection{Analyzing \name's parse decoder}
\label{sec:parse_decoder_analysis}
\noindent\emph{How well does the predicted chunk sequence match the ground truth? } We evaluate the generated chunk sequences by the parse decoder to explore how well it can recover the ground-truth chunk sequence (where the ``ground truth'' is provided by the external parser). Concretely, we compute the chunk-level F1 between the predicted chunk sequence and the ground-truth. We evaluate two configurations of the parse decoder, one in which it is trained separately from the token decoder (first column of Table~\ref{tab:synst_parse_stats}), and the other where both decoders are trained jointly (second column of Table~\ref{tab:synst_parse_stats}). We observe that joint training boosts the chunk F1 from 65.4 to 69.6, although, in both cases the F1 scores are relatively low, which matches our intuition as most source sentences can be translated into multiple target syntactic forms.\smallbreak

\noindent\emph{How much does the token decoder rely on the predicted chunk sequence? }
If \name's token decoder produces the translation ``the man went to the store'' from the parse decoder's prediction of \texttt{PP3 NP3}, it has clearly ignored the predicted chunk sequence. To measure how often the token decoder follows the predicted chunk sequence, we parse the generated translation and compute the F1 between the resulting chunk sequence and the parse decoder's prediction (fourth column of Table~\ref{tab:synst_parse_stats}). Strong results of 89.9 F1 and 43.1\% exact match indicate that the token decoder is heavily reliant on the generated chunk sequences.\smallbreak

\noindent\emph{When the token decoder deviates from the predicted chunk sequence, does it do a better job matching the ground-truth target syntax?} 
Our next experiment investigates why the token decoder sometimes ignores the predicted chunk sequence. One hypothesis is that it does so to correct mistakes made by the parse decoder. To evaluate this hypothesis, we parse the predicted translation (as we did in the previous experiment) and then compute the chunk-level F1 between the resulting chunk sequence and the \emph{ground-truth} chunk sequence. The resulting F1 is indeed almost 10 points higher (third column of Table~\ref{tab:synst_parse_stats}), indicating that the token decoder does have the ability to correct mistakes.\smallbreak

\noindent\emph{What if we vary the max chunk size $k$ during training? } Given a fixed $k$, our chunking algorithm (see Figure~\ref{fig:parse_alg}) produces a deterministic chunking, allowing better control of \name's speedup, even if that sequence may not be optimal for the token decoder. During training we investigate using $k^\prime =\min (k, \sqrt{T})$, where $T$ is the target sentence length (to ensure short inputs do not collapse into a single chunk) and randomly sampling $k\in\{1\ldots6\}$. The final row of Table~\ref{tab:iwslt_bleu_speedup_tradeoff} shows that exposing the parse decoder to multiple possible chunkings of the same sentence during training allows it to choose a sequence of chunks that has a higher likelihood at test time, improving BLEU by 1.5 while decreasing the speedup from $3.8\times$ to $3.1\times$; this is an exciting result for future work (see Table~\ref{tab:synst_parse_stats_extended} for additional analysis).

\section{Related Work}
\label{sec:relwork}

Our work builds on the existing body of literature in both fast decoding methods for neural generation models as well as syntax-based MT; we review each area below.

\subsection{Fast neural decoding}
While all of the prior work described in Section~\ref{sec:background} is relatively recent, non-autoregressive methods for decoding in NMT have been around for longer, although none relies on syntax like \name.~\newcite{schwenk2012continuous} translate short phrases non-autoregressively, while~\newcite{kaiser2016can} implement a non-autoregressive neural GPU architecture and ~\newcite{libovicky2018end2end} explore a CTC approach. \newcite{guo2019nonautoregressive} use phrase tables and word-level adversarial methods to improve upon the NAT model of~\newcite{nonAutoregressiveTransformer}, while \newcite{wang2019nonautoregressive} regularize NAT by introducing similarity and back-translation terms to the training objective.

\subsection{Syntax-based translation} 
There is a rich history of integrating syntax into machine translation systems.~\newcite{wu1997stochastic} pioneered this direction by proposing an inverse transduction grammar for building word aligners.~\newcite{yamada2001syntax} convert an externally-derived source parse tree to a target sentence, the reverse of what we do with \name's parse decoder; later, other variations such as string-to-tree and tree-to-tree translation models followed~\citep{galley2006scalable,cowan2006discriminative}. The Hiero system of~\newcite{chiang2005hierarchical} employs a learned synchronous context free grammar within phrase-based translation, which follow-up work augmented with syntactic supervision~\citep{zollmann2006syntax,marton2008soft,chiang2008online}.  

Syntax took a back seat with the advent of neural MT, as early sequence to sequence models~\citep{sutskever2014sequence,luong2015effective} focused on architectures and optimization.~\newcite{sennrich2016linguistic} demonstrate that augmenting word embeddings with dependency relations helps NMT, while~\newcite{shi2016does} show that NMT systems do not automatically learn subtle syntactic properties.~\newcite{stahlberg2016syntactically} incorporate Hiero's translation grammar into NMT systems with improvements; similar follow-up results~\citep{parseGuidedNMT,eriguchi2017learning} directly motivated this work.

\section{Conclusions \& Future Work}
We propose \name, a variant of the Transformer architecture that achieves decoding speedups by autoregressively generating a constituency chunk sequence before non-autoregressively producing all tokens in the target sentence. Controlled experiments show that \name\ outperforms competing non- and semi-autoregressive approaches in terms of both BLEU and wall-clock speedup on En-De and En-Fr language pairs. While our method is currently restricted to languages that have reliable constituency parsers, an exciting future direction is to explore unsupervised tree induction methods for low-resource target languages~\citep{drozdov2019diora}. Finally, we hope that future work in this area will follow our lead in using carefully-controlled experiments to enable meaningful comparisons.

\section*{Acknowledgements}
\label{sec:acknowledge}

We thank the anonymous reviewers for their insightful comments. We also thank Justin Payan and the rest of the UMass NLP group for helpful comments on earlier drafts. Finally, we thank Weiqiu You for additional experimentation efforts.

\bibliography{bib/journal-full,bib/nsa.bib}

\begin{thebibliography}{36}
\expandafter\ifx\csname natexlab\endcsname\relax\def\natexlab#1{#1}\fi

\bibitem[{Aharoni and Goldberg(2017)}]{parseGuidedNMT}
Roee Aharoni and Yoav Goldberg. 2017.
\newblock Towards string-to-tree neural machine translation.
\newblock In \emph{Proceedings of the Association for Computational
  Linguistics}.

\bibitem[{Bahdanau et~al.(2014)Bahdanau, Cho, and Bengio}]{bahdanau2014neural}
Dzmitry Bahdanau, Kyunghyun Cho, and Yoshua Bengio. 2014.
\newblock Neural machine translation by jointly learning to align and
  translate.
\newblock In \emph{Proceedings of the International Conference on Learning
  Representations}.

\bibitem[{Chiang(2005)}]{chiang2005hierarchical}
David Chiang. 2005.
\newblock A hierarchical phrase-based model for statistical machine
  translation.
\newblock In \emph{Proceedings of the Association for Computational
  Linguistics}, pages 263--270. Association for Computational Linguistics.

\bibitem[{Chiang et~al.(2008)Chiang, Marton, and Resnik}]{chiang2008online}
David Chiang, Yuval Marton, and Philip Resnik. 2008.
\newblock Online large-margin training of syntactic and structural translation
  features.
\newblock In \emph{Proceedings of Empirical Methods in Natural Language
  Processing}, pages 224--233. Association for Computational Linguistics.

\bibitem[{Cowan et~al.(2006)Cowan, Ku{\v{c}}erov{\'a}, and
  Collins}]{cowan2006discriminative}
Brooke Cowan, Ivona Ku{\v{c}}erov{\'a}, and Michael Collins. 2006.
\newblock A discriminative model for tree-to-tree translation.
\newblock In \emph{Proceedings of Empirical Methods in Natural Language
  Processing}, pages 232--241. Association for Computational Linguistics.

\bibitem[{Drozdov et~al.(2019)Drozdov, Verga, Yadav, Iyyer, and
  McCallum}]{drozdov2019diora}
Andrew Drozdov, Patrick Verga, Mohit Yadav, Mohit Iyyer, and Andrew McCallum.
  2019.
\newblock Unsupervised latent tree induction with deep inside-outside recursive
  autoencoders.
\newblock In \emph{Conference of the North American Chapter of the Association
  for Computational Linguistics}.

\bibitem[{Eriguchi et~al.(2017)Eriguchi, Tsuruoka, and
  Cho}]{eriguchi2017learning}
Akiko Eriguchi, Yoshimasa Tsuruoka, and Kyunghyun Cho. 2017.
\newblock Learning to parse and translate improves neural machine translation.
\newblock In \emph{Proceedings of the Association for Computational
  Linguistics}. Association for Computational Linguistics (ACL).

\bibitem[{Galley et~al.(2006)Galley, Graehl, Knight, Marcu, DeNeefe, Wang, and
  Thayer}]{galley2006scalable}
Michel Galley, Jonathan Graehl, Kevin Knight, Daniel Marcu, Steve DeNeefe, Wei
  Wang, and Ignacio Thayer. 2006.
\newblock Scalable inference and training of context-rich syntactic translation
  models.
\newblock In \emph{Proceedings of the Association for Computational
  Linguistics}, pages 961--968. Association for Computational Linguistics.

\bibitem[{Gu et~al.(2018)Gu, Bradbury, Xiong, Li, and
  Socher}]{nonAutoregressiveTransformer}
Jiatao Gu, James Bradbury, Caiming Xiong, Victor~OK Li, and Richard Socher.
  2018.
\newblock Non-autoregressive neural machine translation.
\newblock In \emph{Proceedings of International Conference on Learning
  Representations}.

\bibitem[{Guo et~al.(2019)Guo, Tan, He, Qin, Xu, and
  Liu}]{guo2019nonautoregressive}
Junliang Guo, Xu~Tan, Di~He, Tao Qin, Linli Xu, and Tie-Yan Liu. 2019.
\newblock Non-{Autoregressive} {Neural} {Machine} {Translation} with {Enhanced}
  {Decoder} {Input}.
\newblock In \emph{Association for the Advancement of Artificial Intelligence}.

\bibitem[{Hochreiter and Schmidhuber(1997)}]{hochreiter1997long}
Sepp Hochreiter and J{\"u}rgen Schmidhuber. 1997.
\newblock Long short-term memory.
\newblock \emph{Neural computation}.

\bibitem[{Iyyer et~al.(2018)Iyyer, Wieting, Gimpel, and
  Zettlemoyer}]{IyyerSCPN2018}
Mohit Iyyer, John Wieting, Kevin Gimpel, and Luke Zettlemoyer. 2018.
\newblock Adversarial example generation with syntactically controlled
  paraphrase networks.
\newblock In \emph{Conference of the North American Chapter of the Association
  for Computational Linguistics}.

\bibitem[{Kaiser and Bengio(2016)}]{kaiser2016can}
{\L}ukasz Kaiser and Samy Bengio. 2016.
\newblock Can active memory replace attention?
\newblock In \emph{Proceedings of Advances in Neural Information Processing
  Systems}, pages 3781--3789.

\bibitem[{Kaiser et~al.(2018)Kaiser, Bengio, Roy, Vaswani, Parmar, Uszkoreit,
  and Shazeer}]{latentTransformer}
Lukasz Kaiser, Samy Bengio, Aurko Roy, Ashish Vaswani, Niki Parmar, Jakob
  Uszkoreit, and Noam Shazeer. 2018.
\newblock \href {http://proceedings.mlr.press/v80/kaiser18a.html} {Fast
  decoding in sequence models using discrete latent variables}.
\newblock In \emph{Proceedings of the 35th International Conference on Machine
  Learning}, volume~80 of \emph{Proceedings of Machine Learning Research},
  pages 2390--2399, Stockholmsm{\"a}ssan, Stockholm Sweden. PMLR.

\bibitem[{Kim and Rush(2016)}]{kim2016sequence}
Yoon Kim and Alexander~M Rush. 2016.
\newblock Sequence-level knowledge distillation.
\newblock In \emph{Proceedings of Empirical Methods in Natural Language
  Processing}, pages 1317--1327.

\bibitem[{Lee et~al.(2018)Lee, Mansimov, and Cho}]{lee2018deterministic}
Jason Lee, Elman Mansimov, and Kyunghyun Cho. 2018.
\newblock Deterministic non-autoregressive neural sequence modeling by
  iterative refinement.
\newblock In \emph{Proceedings of Empirical Methods in Natural Language
  Processing}, pages 1173--1182.

\bibitem[{Libovický and Helcl(2018)}]{libovicky2018end2end}
Jindřich Libovický and Jindřich Helcl. 2018.
\newblock End-to-{End} {Non}-{Autoregressive} {Neural} {Machine} {Translation}
  with {Connectionist} {Temporal} {Classification}.
\newblock In \emph{Proceedings of Empirical Methods in Natural Language
  Processing}.

\bibitem[{Luong et~al.(2015)Luong, Pham, and Manning}]{luong2015effective}
Thang Luong, Hieu Pham, and Christopher~D Manning. 2015.
\newblock Effective approaches to attention-based neural machine translation.
\newblock In \emph{Proceedings of Empirical Methods in Natural Language
  Processing}, pages 1412--1421.

\bibitem[{Manning et~al.(2014)Manning, Surdeanu, Bauer, Finkel, Bethard, and
  McClosky}]{corenlp}
Christopher~D. Manning, Mihai Surdeanu, John Bauer, Jenny Finkel, Steven~J.
  Bethard, and David McClosky. 2014.
\newblock \href {http://www.aclweb.org/anthology/P/P14/P14-5010} {The
  {Stanford} {CoreNLP} natural language processing toolkit}.
\newblock In \emph{Association for Computational Linguistics (ACL) System
  Demonstrations}, pages 55--60.

\bibitem[{Marton and Resnik(2008)}]{marton2008soft}
Yuval Marton and Philip Resnik. 2008.
\newblock Soft syntactic constraints for hierarchical phrased-based
  translation.
\newblock \emph{Proceedings of the Association for Computational Linguistics},
  pages 1003--1011.

\bibitem[{Post(2018)}]{sacreBLEU}
Matt Post. 2018.
\newblock \href {http://aclweb.org/anthology/W18-6319} {A call for clarity in
  reporting {BLEU} scores}.
\newblock In \emph{Proceedings of the Third Conference on Machine Translation:
  Research Papers}, pages 186--191. Association for Computational Linguistics.

\bibitem[{Schwenk(2012)}]{schwenk2012continuous}
Holger Schwenk. 2012.
\newblock Continuous space translation models for phrase-based statistical
  machine translation.
\newblock \emph{Proceedings of International Conference on Computational
  Linguistics}, pages 1071--1080.

\bibitem[{Sennrich and Haddow(2016)}]{sennrich2016linguistic}
Rico Sennrich and Barry Haddow. 2016.
\newblock Linguistic input features improve neural machine translation.
\newblock In \emph{Proceedings of the First Conference on Machine Translation:
  Volume 1, Research Papers}, volume~1, pages 83--91.

\bibitem[{Sennrich et~al.(2016)Sennrich, Haddow, and Birch}]{bpe}
Rico Sennrich, Barry Haddow, and Alexandra Birch. 2016.
\newblock Neural machine translation of rare words with subword units.
\newblock In \emph{Proceedings of the 54th Annual Meeting of the Association
  for Computational Linguistics (Volume 1: Long Papers)}, volume~1, pages
  1715--1725.

\bibitem[{Shi et~al.(2016)Shi, Padhi, and Knight}]{shi2016does}
Xing Shi, Inkit Padhi, and Kevin Knight. 2016.
\newblock Does string-based neural mt learn source syntax?
\newblock In \emph{Proceedings of Empirical Methods in Natural Language
  Processing}, pages 1526--1534.

\bibitem[{Stahlberg et~al.(2016)Stahlberg, Hasler, Waite, and
  Byrne}]{stahlberg2016syntactically}
F~Stahlberg, E~Hasler, A~Waite, and B~Byrne. 2016.
\newblock Syntactically guided neural machine translation.
\newblock In \emph{Proceedings of the Association for Computational
  Linguistics}, volume~2, pages 299--305. Association for Computational
  Linguistics.

\bibitem[{Stern et~al.(2018)Stern, Shazeer, and Uszkoreit}]{parallelDecoding}
Mitchell Stern, Noam Shazeer, and Jakob Uszkoreit. 2018.
\newblock Blockwise parallel decoding for deep autoregressive models.
\newblock In \emph{Advances in Neural Information Processing Systems}, pages
  10106--10115.

\bibitem[{Strubell et~al.(2018)Strubell, Verga, Andor, Weiss, and
  McCallum}]{emnlp18-strubell}
Emma Strubell, Patrick Verga, Daniel Andor, David Weiss, and Andrew McCallum.
  2018.
\newblock { Linguistically-Informed Self-Attention for Semantic Role Labeling}.
\newblock In \emph{Proceedings of Empirical Methods in Natural Language
  Processing}, Brussels, Belgium.

\bibitem[{Sutskever et~al.(2014)Sutskever, Vinyals, and
  Le}]{sutskever2014sequence}
Ilya Sutskever, Oriol Vinyals, and Quoc~V Le. 2014.
\newblock Sequence to sequence learning with neural networks.
\newblock In \emph{Proceedings of Advances in Neural Information Processing
  Systems}, pages 3104--3112.

\bibitem[{Swayamdipta et~al.(2018)Swayamdipta, Thomson, Lee, Zettlemoyer, Dyer,
  and Smith}]{swayamdipta2018syntactic}
Swabha Swayamdipta, Sam Thomson, Kenton Lee, Luke Zettlemoyer, Chris Dyer, and
  Noah~A Smith. 2018.
\newblock Syntactic scaffolds for semantic structures.
\newblock In \emph{Proceedings of Empirical Methods in Natural Language
  Processing}, pages 3772--3782.

\bibitem[{Vaswani et~al.(2017)Vaswani, Shazeer, Parmar, Uszkoreit, Jones,
  Gomez, Kaiser, and Polosukhin}]{transformer}
Ashish Vaswani, Noam Shazeer, Niki Parmar, Jakob Uszkoreit, Llion Jones,
  Aidan~N Gomez, {\L}ukasz Kaiser, and Illia Polosukhin. 2017.
\newblock Attention is all you need.
\newblock In \emph{Advances in Neural Information Processing Systems}, pages
  5998--6008.

\bibitem[{Wang et~al.(2018)Wang, Zhang, and
  Chen}]{semiAutoregressiveTransformer}
Chunqi Wang, Ji~Zhang, and Haiqing Chen. 2018.
\newblock Semi-autoregressive neural machine translation.
\newblock In \emph{Proceedings of the 2018 Conference on Empirical Methods in
  Natural Language Processing}, pages 479--488.

\bibitem[{Wang et~al.(2019)Wang, Tian, He, Qin, Zhai, and
  Liu}]{wang2019nonautoregressive}
Yiren Wang, Fei Tian, Di~He, Tao Qin, ChengXiang Zhai, and Tie-Yan Liu. 2019.
\newblock Non-{Autoregressive} {Machine} {Translation} with {Auxiliary}
  {Regularization}.
\newblock In \emph{Association for the Advancement of Artificial Intelligence}.

\bibitem[{Wu(1997)}]{wu1997stochastic}
Dekai Wu. 1997.
\newblock Stochastic inversion transduction grammars and bilingual parsing of
  parallel corpora.
\newblock \emph{Computational linguistics}, 23(3):377--403.

\bibitem[{Yamada and Knight(2001)}]{yamada2001syntax}
Kenji Yamada and Kevin Knight. 2001.
\newblock A syntax-based statistical translation model.
\newblock In \emph{Proceedings of the Association for Computational
  Linguistics}.

\bibitem[{Zollmann and Venugopal(2006)}]{zollmann2006syntax}
Andreas Zollmann and Ashish Venugopal. 2006.
\newblock Syntax augmented machine translation via chart parsing.
\newblock In \emph{Proceedings of the Workshop on Statistical Machine
  Translation}, pages 138--141. Association for Computational Linguistics.

\end{thebibliography}
\bibliographystyle{style/acl_natbib}

\newpage

\setcounter{table}{0} \renewcommand{\thetable}{A\arabic{table}}
\setcounter{figure}{0} \renewcommand{\thefigure}{A\arabic{figure}}

\appendix
\section*{Appendix}

\section{Unscientific Comparison}
\label{appendix:unscientific-comparison}

We include a reference to previously published work in comparison to our approach. Note, that many of these papers have multiple confounding factors that make direct comparison between approaches very difficult.

\begin{table}[ht]
\begin{center}
\begin{tabular}{ lcc }
 \toprule
 \textbf{Model} & \multicolumn{2}{c}{\underline{\textbf{WMT En-De}}} \\
 & BLEU & Speedup \\
 \midrule
 LT rescoring top-100 & 22.5 & - \\
 NAT rescoring top-100 & 21.54 & - \\
 BPT ($k=6$) & 28.11 & $3.10\times$ \\
 IRT (adaptive) & 21.54 & $2.39\times$ \\
 SAT ($k=6$) & 23.93 & $5.58\times$ \\
 \name ($k=6$) & 20.74 & $4.86\times$ \\
 \bottomrule
\end{tabular}
\end{center}
\caption{Unscientific comparison against previously published works. The numbers of each model are taken from their respective papers. These previous results often have uncomparable hyperparameters, compute their BLEU with \texttt{multi-bleu.perl}, and/or require additional steps such as knowledge distillation and re-ranking to achieve their reported numbers. Latent Transformer (LT) \cite{latentTransformer}, Non-autoregressive Transformer (NAT) \cite{nonAutoregressiveTransformer}, Blockwise parallel Transformer (BPT) \cite{parallelDecoding}, Iterative refinement Transformer (IRT) \cite{lee2018deterministic}, Semi-autoregressive Transformer (SAT) \cite{semiAutoregressiveTransformer}.}
\label{tab:experiments-uncontrolled}
\end{table}

\section{The impact of beam search}

In order to more fully understand the interplay of the representations output from the autoregressive parse decoder on the BLEU/speedup tradeoff we examine the impact of beam search for the parse decoder. From Table~\ref{tab:beam_search_impact} we see that beam search does not consitently improve the final translation quality in terms of BLEU (it manages to decrease BLEU on IWSLT), while providing a small reduction in overall speedup for \name.

\section{SAT replication results}

As part of our work, we additionally replicated the results of \citep{semiAutoregressiveTransformer}. We do so without any of the additional training stabilization techniques they use, such as knowledge distillation or initializing from a pre-trained Transformer. Without the use of these techniques, we notice that the approach sometimes catastrophically fails to converge to a meaningful representation, leading to sub-optimal translation performance, despite achieving adequate perplexity. In order to report accurate translation performance for SAT, we needed to re-train the model for $k=4$ when it produced BLEU scores in the single digits.

\begin{table*}[t]
\begin{center}
\begin{adjustbox}{max width=\textwidth}
\begin{tabular}{ lccccccccc }
 \toprule
 \textbf{Model} & \textbf{Beam} & \multicolumn{2}{c}{\underline{\textbf{WMT En-De}}} & \multicolumn{2}{c}{\underline{\textbf{WMT De-En}}}  & \multicolumn{2}{c}{\underline{\textbf{IWSLT En-De}}} & \multicolumn{2}{c}{\underline{\textbf{WMT En-Fr}}}\\
                & \textbf{Width} & BLEU & Speedup & BLEU & Speedup & BLEU & Speedup & BLEU & Speedup \\
 \midrule
 Transformer & 1 & 25.82 & $1.15\times$ & 29.83 & $1.14\times$  & 28.66 & $1.16\times$ & 39.41 & $1.18\times$ \\
 Transformer & 4 & 26.87 & $1.00\times$ & 30.73 & $1.00\times$  & 30.00 & $1.00\times$ & 40.22 & $1.00\times$ \\
 \hline
 SAT ($k=2$) & 1 & 22.81 & $2.05\times$ & 26.78 & $2.04\times$ & 25.48 & $2.03\times$ & 36.62 & $2.14\times$ \\
 SAT ($k=2$) & 4 & 23.86 & $1.80\times$ & 27.27 & $1.82\times$ & 26.25 & $1.82\times$ & 37.07 & $1.89\times$ \\
 SAT ($k=4$) & 1 & 16.44 & $3.61\times$ & 21.27 & $3.58\times$ & 20.25 & $3.45\times$ & 28.07 & $3.34\times$ \\
 SAT ($k=4$) & 4 & 18.95 & $3.25\times$ & 23.20 & $3.19\times$ & 20.75 & $2.97\times$ & 32.62 & $3.08\times$ \\
 SAT ($k=6$) & 1 & 12.55 & $4.86\times$ & 15.23 & $4.27\times$ & 14.02 & $4.39\times$ & 24.63 & $4.77\times$ \\
 SAT ($k=6$) & 4 & 14.99 & $4.15\times$ & 19.51 & $3.89\times$ & 15.51 & $3.78\times$ & 28.16 & $4.19\times$ \\
 \hline
 \lt*  & - & 19.8 & $3.89\times$ & - & - & - & - & - & - \\
 \hline
 \name ($k=6$) & 1 & 20.74 & $4.86\times$ & 25.50 & $5.06\times$ & 23.82 & $3.78\times$ & 33.47 & $5.32\times$ \\
 \name ($k=6$) & 4 & 21.61 & $3.89\times$ & 25.77 & $4.07\times$ & 23.31 & $3.11\times$ & 34.10 & $4.47\times$ \\
 \bottomrule
\end{tabular}
\end{adjustbox}
\end{center}
\caption{Controlled experiments comparing \name\ to \lt\ and SAT on four different datasets (two language pairs) demonstrate speed and BLEU improvements while varying beam size. Wall-clock speedup is measured on a single Nvidia TitanX Pascal by computing the average time taken to decode a single sentence in the dev/test set, averaged over five runs. Note that the LT results are reported by~\newcite{latentTransformer} and not from our own implementation; as such, they are not directly comparable to the other results.}
\label{tab:beam_search_impact}
\end{table*}

\section{Parse performance when varying max chunk size $k$}

In Section~\ref{sec:parse_decoder_analysis} (see the final row of Table~\ref{tab:synst_parse_stats}) we consider the effect of randomly sampling the max chunk size $k$ during training. This provides a considerable boost to BLEU with a minimal impact to speedup. In Table~\ref{tab:synst_parse_stats_extended} we highlight the impact to the parse decoder's ability to predict the ground-truth chunk sequences and how faithfully it follows the predicted sequence.

\begin{table*}[t]
  \centering
  \begin{tabular}{lcccc}
    \toprule &
    \bf \thead{Max Chunk Size} &
    \bf \thead{Predicted parse\\vs.\\Gold parse} &
    \bf \thead{Parsed prediction\\vs.\\Gold parse} &
    \bf \thead{Parsed prediction\\vs.\\Predicted parse} \\
    \midrule
    F1 & \multirow{2}{*}{$k=6$} & $69.64$ & $79.16$ & $89.90$ \\
    Exact match & & $5.24\%$ & $5.94\%$ & $43.10\%$ \\
    \midrule
    F1 & \multirow{2}{*}{$k\in\{1\ldots6\}$} & $75.35$ & $79.78$ & $95.28$ \\
    Exact match & & $4.83\%$ & $7.55\%$ & $50.15\%$ \\
    \bottomrule
  \end{tabular}
  \caption{F1 and exact match comparisons of predicted chunk sequences (from the parse decoder), ground-truth chunk sequences (from an external parser in the target language), and chunk sequences obtained after parsing the translation produced by the token decoder. The first column shows how well the parse decoder is able to predict the ground-truth chunk sequence when trained jointly with the token decoder. The second column shows that when the token decoder deviates from the predicted chunk sequence, it usually results in a translation that is closer to the ground-truth target syntax, while the third column shows that the token decoder closely follows the predicted chunk sequence. Randomly sampling $k$ from $\{1\ldots6\}$ during training significantly boosts the parse decoder's ability to recover the ground-truth chunk sequence compared to using a fixed $k=6$. Subsequently the token decoder follows the chunk sequence more faithfully.}
  \label{tab:synst_parse_stats_extended}
\end{table*}

\end{document}